\documentclass[journal]{IEEEtran}
\ifCLASSINFOpdf
\else
\fi
\usepackage{graphicx}
\usepackage{epstopdf}
\usepackage{subfigure}
\usepackage{amssymb,amsmath}
\usepackage{amsfonts}
\usepackage[ruled]{algorithm2e}
\usepackage{algpseudocode}
\usepackage{multirow}
\usepackage{multicol}
\usepackage[flushleft]{threeparttable}
\usepackage{balance}

\begin{document}
%

\title{Transductive Zero-Shot Learning with a Self-training dictionary approach}
%
%
%

\author{Yunlong~Yu,~Zhong~Ji,~Xi~Li,~Jichang~Guo,~Zhongfei~Zhang, Haibin Ling, Fei Wu


}

\maketitle

\begin{abstract}
    As an important and challenging problem in computer vision, zero-shot learning (ZSL) aims at automatically recognizing the instances from unseen object classes without training data. To address this problem, ZSL is usually carried out in the following two aspects: 1) capturing the domain distribution connections between seen classes data and unseen classes data; and 2) modeling the semantic interactions between the image feature space and the label embedding space. Motivated by these observations, we propose a bidirectional mapping based semantic relationship modeling scheme that seeks for cross-modal knowledge transfer by simultaneously projecting the image features and label embeddings into a common latent space. Namely, we have a bidirectional connection relationship that takes place from the image feature space to the latent space as well as from the label embedding space to the latent space. To deal with the domain shift problem, we further present a transductive learning approach that formulates the class prediction problem in an iterative refining process, where the object classification capacity is progressively reinforced through bootstrapping-based model updating over highly reliable instances. Experimental results on three benchmark datasets (AwA, CUB and SUN) demonstrate the effectiveness of the proposed approach against the state-of-the-art approaches.

\end{abstract}


%
\IEEEpeerreviewmaketitle

\section{Introduction}
%
%
%
%

    \IEEEPARstart{Z}ero-shot learning (ZSL) \cite{IEEEcvpr09:Lampert,IEEEcvpr15:Akata,IEEEcvpr16:Xian,IEEEcvpr15:Fu,IEEEcvpr13:Akata,ICLR14:Norouzi,IEEEnips15:Romera-Paredes} endows the computer vision system with the capability to recognize instances of a new class that has never seen before. A common framework to address this problem is to transfer the knowledge from the seen classes to unseen ones by resorting to a label embedding space where the semantic relatedness between different classes are measured. Commonly used semantic label embeddings include visual attributes \cite{IEEEcvpr09:Lampert, IEEEcvpr15:Akata, IEEEcvpr13:Akata, IEEEcvpr16:Changpinyo, IEEEeccv14:Fu, IEEEpami:Lampert} and word vectors \cite{IEEEcvpr15:Akata, IEEEcvpr16:Xian, IEEEnips13:Frome}.

    In order to achieve the knowledge transfer, existing approaches fall into two main categories. The first one poses the seen classes as the mediators to connect the test instance and the unseen classes. It relies on learning a classification model for seen classes with the labeled instances, which is then used to compute the visual similarities between the test instance and seen classes. The prediction is implemented by matching the visual similarities and the semantic relatedness between the seen classes and the unseen classes, which is obtained with their label embeddings. In contrast, the approaches in the second category focus on modeling the semantic interactions between different modalities by directly learning a projection function either from the image feature space to the label embedding space \cite{IEEEnips13:Socher, icip15:Xu}, or from a reverse direction \cite{IEEEiccv15:Kodirov, arXiv16:Shojaee}, and then predict the unseen instances in the label embedding space or image visual space.

    A common characteristic of existing ZSL approaches from both categories is that they all critically rely on the pre-defined label embeddings to compute the semantic relatedness between the seen and unseen classes. However, the noisy and uncertainty of the label embedding make it hard to characterize the semantic information explicitly, which will be blindly forced to the unseen data during the knowledge transfer. Besides, we only have a single sparse label semantic vector for each unseen class, which is insufficient to fully represent the data distribution of the class. Thus, the distribution connections between the seen domain and unseen domain are difficult to capture. Motivated by these observations, we propose a bidirectional mapping based semantic relationship modeling scheme that seeks for cross-modal knowledge transfer by simultaneously projecting the image features and label embeddings into a common latent space. In specific, the bidirectional connection relationship is formulated into a general dictionary framework, in which a common latent space is learned for preserving the semantic relatedness between different modalities. By projecting the label embeddings to the latent space where the embedding semantics are more suitably aligned, the influence of semantic gap across different modalities alleviates.

    As the seen classes and unseen classes are different and potentially unrelated, the projection function learned from the seen domain is usually biased on the unseen domain. To address this domain shift issue, many approaches focus on learning a more general projection function to bridge the semantic relationships between the image feature space and the label embedding space under a transductive setting \cite{arXiv16:Shojaee, IEEEaaai16:Guo,arXiv16:Wang,IEEEpami15:Fu}. The transductive setting means that the unlabeled unseen instances are used to improve the generalization accuracy. However, existing transductive approaches treat all unlabeled data equally and achieve the prediction in one pass, which makes the learned models difficult to relate the seen domain to the unseen domain. Based on this motivation, we further present a transductive learning approach that treats the unlabeled unseen instances in different levels by assessing their reliability and discriminability. Specifically, it formulates the class prediction problem in an iterative refining process, in which each iteration alternates between two paradigms, learning-to-predict and predicting-to-learn. In the learning-to-predict paradigm, the prediction is conducted on the unseen data with the current learned model to select reliable instances for the subsequent learning process; In the predicting-to-learn paradigm, the model is retrained with the feedback reliable instances for the next prediction. In this way, the object classification capacity is progressively reinforced through bootstrapping-based model updating over highly reliable instances.

    The flowchart of the proposed transductive ZSL approach is illustrated in Fig.~1. In conclusion, the main contributions of this paper are two folds:\\

    \begin{enumerate}
      \item  To achieve the knowledge transfer from the seen classes data to the unseen classes data, we propose a general dictionary model to simultaneously project the image features and label embeddings into a common latent space, where the class semantic relatedness between different modalities are effectively preserved.
      \item A novel transductive framework is developed for alleviating the domain shift problem in ZSL by formulating the class prediction step in an iterative refining process, in which the domain shift is gradually adapted by retraining a powerful classification model with highly reliable unseen instances. Experimental results show that the proposed transductive strategy can significantly improve the inductive classification model and outperform the state-of-the-art related approaches.
    \end{enumerate}


 \begin{figure}
  \centering
  \includegraphics[height=6.5cm,width=8.5cm]{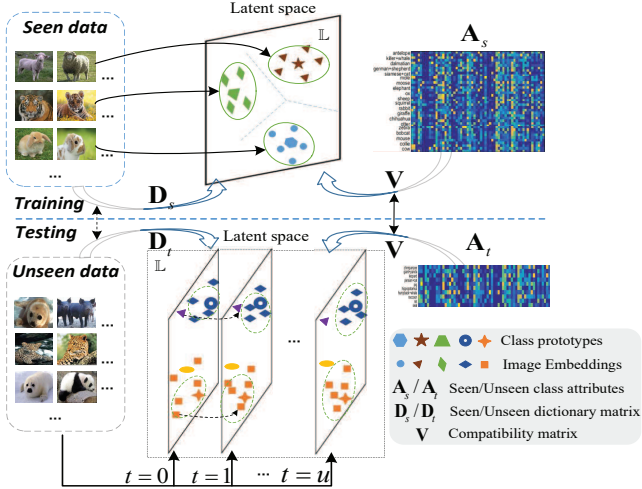}
  \caption{The illustration of our proposed model with attributes. In the training stage, the visual image features and the class attribute features are jointly embedded in the latent space, where the compatibility scores between different domains are obtained. In the testing stage, the previous predicted results are obtained with the learned dictionary matrix $\mathbf{D}_s$ from the seen domain and the compatibility matrix $\mathbf{V}$, and the predicted results are updated in an iterative refining process. At each iteration, the prototypes of the unseen classes are fixed in the latent space, and the unseen instances with high compatibility scores are selected for retraining a more powerful dictionary model $\mathbf{D}_t$ for unseen classes. $u$ is the number of iteration.}
  \label{Fig.1}
 \end{figure}

    The remaining sections are organized as follows. Section \uppercase\expandafter{\romannumeral2} describes the related work. Section \uppercase\expandafter{\romannumeral3} presents the proposed general dictionary model for achieving the cross-modal knowledge transfer and the transductive framework for addressing the domain shift problem in ZSL. Section \uppercase\expandafter{\romannumeral4} provides extensive experiments and evaluations, followed by the conclusion in Section \uppercase\expandafter{\romannumeral5}.

\section{Related work}
\subsection{Knowledge transfer for ZSL}
    The key idea of ZSL is transferring the knowledge from the seen domain to the unseen one. It relies on constructing a label semantic embedding space where each class can be represented as a vector and the semantic relationships among all classes can be precisely characterized. The most common label embeddings include visual attributes \cite{IEEEcvpr09:Lampert}, \cite{IEEEcvpr13:Akata}, \cite{cvpr13:Yu}, word vectors \cite{IEEEcvpr16:Qiao}, \cite{IEEEnips13:Frome}, \cite{cvpr16:Halah}, knowledge mined from the Web \cite{cvpr10:Rohrbach}, \cite{cvpr14:Mensink}. Visual attributes are a list of manually specified properties for categories, such as color, shape and presence or absence of a certain body part, which are shared across both the seen and unseen classes. In contrast, semantic word vectors are obtained from a large text corpus in an unsupervised way. With a language model, such as word2vec \cite{IEEEnips13:Mikolov} and Glove \cite{IEEEemnl14:Pennington}, each class name is embedded into the word vector space, where the class semantic information is defined. Given such label semantic embeddings, the existing approaches of ZSL focus on bridging the class semantic relationships between the instances and the categories with the help of label semantic embeddings. One of the pioneering studies is \cite{IEEEcvpr09:Lampert}, in which two probabilistic paradigms are proposed, i.e., directed attribute prediction (DAP) and indirected attribute prediction (IAP). DAP takes advantage of the class attributes as the middle layer between the input images and the output class labels. However, in IAP model, the seen classes are taken as the middle layer to connect the visual instances and the unseen classes, where the semantic relationships between seen classes and unseen classes are defined by their corresponding attributes.

    Considering that the visual instances and label embeddings are embedded in different spaces, recent work addresses ZSL by exploring the semantic relationships between the visual instances and the label embeddings, which has been widely explored in two ways: (1) learning a direct projection function by regressing from image feature space to the label embedding space with regressors \cite{IEEEcvpr09:Farhadi} \cite{IEEEpami:Lampert} or neural networks \cite{IEEEnips13:Socher}; (2) projecting the visual features and label embeddings into a latent space, such as CCA \cite{IEEEeccv14:Fu}. Instead of learning two different mapping functions for image feature space and label embedding space, SJE \cite{IEEEcvpr15:Akata} and DeViSE \cite{IEEEnips13:Frome} combined the visual features and label embeddings into a bilinear model to represent the compatibility scores of different modalities and employ a ranking objective to enforce the correct class labels to be ranked higher than any of the other class labels. In order to improve the compatibility learning framework, \cite{IEEEcvpr16:Xian} introduced a list of latent variables to learn a collection of mappings with the selection of the latent variable to match the current image-class pair. Taking the class labels into consideration, \cite{IEEEnips15:Romera-Paredes} proposed a simpler but more efficient method that associates the visual feature, label embedding and class label into an joint model. As an extension of \cite{IEEEnips15:Romera-Paredes}, Qiao \emph{et al}. \cite{IEEEcvpr16:Qiao} proposed an $\ell_{2,1}$-norm based objective function which can simultaneously suppress the noisy signal in the textual representation and learn a function to match the textual semantic vectors and visual features.

    Instead of projecting the visual features into the label embedding space, \cite{ECML15:Shigeto} showed that mapping label semantic vectors into the image feature space is desirable to suppress the emergence of hubs in the subsequent nearest neighbor search step. Analogously, \cite{IEEEiccv15:Kodirov} employed a dictionary learning scheme in which class attributes are considered to be coding coefficients which are used to reconstruct the visual instances. Based on the dictionary learning, Zhang \emph{et al.} \cite{IEEEcvpr16:Zhang} proposed a latent probabilistic model to simultaneously project both the visual features and label embeddings into different latent spaces, and then learn a cross-domain similarity matrix for matching different modalities.

\subsection{Adaptation for domain shift problem}
    Domain shift problem is a common issue in the situations where there are a lot training data in one domain but little to none in another. Traditional domain adaptation approaches are derived for both with \cite{cvpr09:Duan}, \cite{ICML:11:Glorot} and without \cite{iccv13:Fernando} requiring label information of the target domain. Since the label information of the unseen domain are not available in ZSL, thus the supervised domain adaptation approaches are not applicable for ZSL. Besides, different from the traditional domain shift problem \cite{cybernetics16:Uzair}, \cite{tnnls12:Duan}, the domain shift issue in ZSL is mainly due to the disjointness of the seen classes and unseen classes rather than the feature distribution shift. Recently, several work has proposed for mitigating domain shift problem in ZSL with methods ranging from subspace aligning \cite{IEEEeccv14:Fu}, data augmentation \cite{IEEEkde10:Pan}, \cite{IEEEnnls15:Shao}, self-training \cite{arXiv15:Xu} to hubness correction \cite{ICLR15:Dinu}. Transductive zero-shot learning was first considered by Fu \emph{et al.} \cite{eccv12:Fu}, in which the unseen data attribute distribution is exploited by averaging the label prototype's k-nearest neighbours. In \cite{IEEEpami15:Fu}, the domain shift problem was addressed by transductive multi-view hypergraph label propagation (TMV-HLP), in which the manifold structure of the unseen data is exploited to compensate for the the impoverished supervision available from the sparse semantic vector. By using graph-based label propagation to exploit the manifold structure of the unseen data, Rohrbach \emph{et al.} \cite{nips13:Rohrbach} proposed a more elaborate transductive strategy for domain shift problem in ZSL. Different from these approaches, Xu \emph{et al.} \cite{icip15:Xu} proposed a data augmentation strategy by mitigating any available auxiliary dataset to the labeled seen data for training a general model for unseen data. Self-training adaptation \cite{icip15:Xu, arXiv16:Wang} was a post-processing technique, which is based on adjusting the latent embeddings of unseen classes according to the distribution of all the test instance projections in the latent subspace.

\section{The proposed model}

    In this section, we focus on learning a specific classification model for recognizing the unlabeled unseen data. It consists of two parts: i) a general dictionary model is learned with the labeled seen data for initially predicting the unseen data, in which the semantic relatedness between different modalities are preserved by projecting the image features and label embeddings into a common latent space; ii) a transductive framework is presented for mitigating domain shift problem in ZSL by formulating the prediction step in an iterative refining precess, where the classification capacity is progressively reinforced through bootstrapping-based model updating over highly reliable instances.

\subsection{Notations}
    Suppose that we collect $m$ labeled instances from $M$ seen classes for training, and each class is associated with a vector embedded in the label embedding space. Denote $\mathbf{X}_{s}=[\mathbf{x}_1^{s},...,\mathbf{x}_m^s]\in\mathbb{R}^{{p}\times{m}}$ as the instances available at training stage, where $p$ is the dimensionality of the image feature. And we use $\mathbf{Y}_{s}\in{\{-1,1\}}^{{m}\times{M}}$ and $\mathbf{A}_{s}\in{\mathbb{R}^{{q}\times{M}}}$ to denote the corresponding ground truth label matrix and label embedding matrix for seen data, respectively. Here, each column of $\mathbf{A}_{s}$ represents a label embedding vector, and $q$ is the dimensionality of the class label embedding. At testing stage, we are given $n$ instances $\mathbf{X}_{t}=[\mathbf{x}_1^t,...,\mathbf{x}_n^t]\in\mathbb{R}^{{p}\times{n}}$ from $N$ unseen classes, which are disjoint from seen classes. Each unseen class is also associated with a label embedding vector. TABLE \uppercase\expandafter{\romannumeral1} shows the main notations used here in after.

    \begin{table}
    \label{Table.1}
    \caption{\upshape The notations used in our approach.}
    \centering
    \begin{tabular}{c|c}
    \hline
    Notation & Description\\
    \hline
    $M$  & number of the seen classes\\
    $N$  & number of the unseen classes\\
    $m$  & number of the seen instances\\
    $n$  & number of the unseen instances\\
    $p$  & dimensionality of the image feature space\\
    $q$  & dimensionality of the label embedding space\\
    $d$  & dimensionality of the latent space\\
    $\mathbf{X}_{s}\in\mathbb{R}^{{p}\times{m}}$ & seen instance matrix\\
    $\mathbf{A}_{s}\in{\mathbb{R}^{{q}\times{M}}}$ & label semantic matrix of seen classes\\
    $\mathbf{Y}_{s}\in{\{-1,1\}}^{{m}\times{M}}$ & ground truth label matrix of seen classes\\
    $\mathbf{X}_{t}\in\mathbb{R}^{{p}\times{n}}$ & unseen instance matrix\\
    $\mathbf{A}_{t}\in{\mathbb{R}^{{q}\times{N}}}$ & label semantic matrix of unseen classes\\
    $\mathbf{V}\in{\mathbb{R}^{{d}\times{q}}}$ & shared compatibility matrix\\ 
    $\mathbf{D}_{s}\in{\mathbb{R}^{{p}\times{d}}}$ & dictionary model for seen data\\
    $\mathbf{D}_{t}\in{\mathbb{R}^{{p}\times{d}}}$ & dictionary model for unseen data\\
    $\alpha$,~$\beta$,~$\lambda$,~$\mu$ & hyper-parameters\\
    $\delta$ & self-labeled rate\\
    $n_i$ & the predicted instance number of $i$-th unseen class\\
    \hline
    \end{tabular}
    \end{table}
\subsection{The Joint Embedding Dictionary Model (JEDM)}
    For the labeled seen data, conventional dictionary learning models \cite{IEEEpami13:Jiang,IEEEpami16:Wang,IEEEnips14:Gu,cybernetic16:Song} aim at learning an effective data representation model from the input data $\mathbf{X}_{s}$ for classification tasks by exploiting the class label discriminative information of labeled data.
    Most existing dictionary learning approaches can be formulated under the following framework:
    \begin{equation}
    \label{Eq.(1)}
    \begin{aligned}
    \{\mathbf{D}_{s}^*,\mathbf{C}_{s}^*,\mathbf{W}^*\}=\arg\min_{\footnotesize{\mathbf{D}_{s},\mathbf{C}_{s},\mathbf{W}}}
    \|\mathbf{X}_{s}-\mathbf{D}_{s}\mathbf{C}_{s}\|_F^2\\
    +\lambda\|\mathbf{C}_{s}\|_p
    +\Psi(\mathbf{W},\mathbf{C}_{s},\mathbf{Y}_{s}),
    \end{aligned}
    \end{equation}
    where $\lambda$ is a weight parameter, $\mathbf{Y}_s$ denotes the class label matrix of the instances from input data matrix $\mathbf{X}_s$, $\mathbf{D}_s$ is the dictionary matrix to be learned, and $\mathbf{C}_s$ is the representative coding matrix of $\mathbf{X}_s$ over $\mathbf{D}_s$, and $\mathbf{W}=[\mathbf{w}_1,...,\mathbf{w}_M]$
     is the classification matrix for seen classes, $\mathbf{w}_i$ is the classifier for class $i$, $1\leq{i}\leq{M}$; $\|\mathbf{C}_{s}\|_p$ is the $\ell_p$-norm regularizer on $\mathbf{C}_{s}$, $\|\mathbf{X}_{s}-\mathbf{D}_{s}\mathbf{C}_{s}\|_F^2$ is the reconstruction error term ensuring the representative ability of $\mathbf{D}_{s}$, $\|\cdot\|_F$ denotes the matrix Frobenious norm, and $\Psi(\mathbf{W},\mathbf{C}_{s},\mathbf{Y}_{s})$ is a discriminative function, which ensures the discriminative ability of $\mathbf{C}_{s}$.

    With Eq.~(1), the shared dictionary matrix and classification matrix can be trained with labeled seen classes. However, no labeled data are available for unseen classes such that the classification parameters for unseen classes cannot be obtained directly. We thereby need to transfer the knowledge exploited from the labeled seen domain to the unseen domain. As previous work has indicated that the properties of a class can be well characterized by its corresponding label embedding, thus it is reasonable to assume that the classifier of a class can be derived from its label embedding. Thus, we replace the classification model $\mathbf{w}$ with: $\mathbf{V}\mathbf{a}$, where $\mathbf{a}$ is the label embedding and $\mathbf{V}$ is the compatibility matrix shared both the seen and unseen classes. Intuitively, the compatibility matrix aligns the semantic consistency between the visual instances and the label embeddings. Once obtaining the compatibility matrix $\mathbf{V}$, the classification parameter $\mathbf{w}_i$ for unseen class $i$ can be obtained by $\mathbf{w}_i = \mathbf{V}\mathbf{a}_i$. To this end, the remaining problem is to learn the compatibility matrix with the labeled seen data. Based on this idea, we propose to learn such a compatibility matrix together with the seen dictionary matrix. Formally, we get the Joint Embedding Dictionary Model (JEDM) for ZSL,
    \begin{equation}
    \label{Eq.(2)}
    \begin{aligned}
    \{\mathbf{D}_{s}^*,\mathbf{C}_{s}^*,\mathbf{V}^*\}=\arg\min_{\footnotesize{\mathbf{D}_{s},\mathbf{C}_{s},\mathbf{V}}}
    \|\mathbf{X}_{s}-\mathbf{D}_{s}\mathbf{C}_{s}\|_F^2\\
    +\alpha\|\mathbf{C}_{s}^\textrm{T}\mathbf{V}\mathbf{A}_{s}-\mathbf{Y}_{s}\|_F^2+\beta\|\mathbf{V}\mathbf{A}_{s}\|_F^2,~ \|d_{i}\|_2^2\leqslant{1},
    \end{aligned}
    \end{equation}
    where $\alpha$ and $\beta$ are two parameters to trade-off different terms, which can be determined via the cross-validation.

    The first term of Eq.~(2) is the reconstruction error, which compresses the visual features in a more representative latent space, and the second term incorporates the latent features, label embeddings and class labels into a joint framework for preserving the semantic relatedness across different modalities. By enforcing the visual latent features being close to the corresponding label embeddings while be far away from that of the other classes, this term is subject to exploit the semantic discriminant information across different modalities. The last term is a regularizer term.


    We next introduce the optimization process to solve the objective function in Eq.~(2). Eq.~(2) is not convex for $\mathbf{D}_{s}$, $\mathbf{C}_{s}$ and $\mathbf{V}$ simultaneously but is convex for each of them individually. Therefore, the optimization can be done alternatively between the following two steps.

    1). Fix $\mathbf{D}_{s}$, $\mathbf{V}$ and solve for $\mathbf{C}_{s}$.
    \begin{equation}
    \begin{aligned}
    \label{Eq.(3)}
    \mathbf{C}_{s}^*=\arg\min_{\mathbf{C}_{s}}
    \|\mathbf{X}-\mathbf{D}_{s}\mathbf{C}_{s}\|_F^2+\alpha\|\mathbf{C}_{s}^\textrm{T}\mathbf{V}\mathbf{A}_{s}-\mathbf{Y}_{s}\|_F^2.
    \end{aligned}
    \end{equation}

    This sub-problem is a standard least square problem; so we take the derivative of Eq.~(3) with respect to $\mathbf{C}_{s}$ and make it equal to zero, which has the following closed-form solution:
    \begin{equation}
    \begin{aligned}
    \mathbf{C}_{s}^*= (\mathbf{D}_{s}^\textrm{T}\mathbf{D}_{s}+\alpha\mathbf{V}\mathbf{A}_{s}\mathbf{A}_{s}^\textrm{T}\mathbf{V}^\textrm{T})^{-1}(\mathbf{D}_{s}^\textrm{T}\mathbf{X}_{s}+
    \alpha\mathbf{V}\mathbf{A}_{s}\mathbf{Y}_{s}^\textrm{T}).
    \end{aligned}
    \end{equation}
\par\setlength\parindent{1em}
    2). Fix $\mathbf{C}_{s}$ and solve $\mathbf{V}$ and $\mathbf{D}_{s}$.
    Since $\mathbf{V}$ and $\mathbf{D}_{s}$ are independent, thus they can be solved separately,
    \begin{equation}
    \mathbf{V}^*= \arg\min_{\mathbf{V}}\alpha\|\mathbf{C}_{s}^\textrm{T}\mathbf{V}\mathbf{A}_{s}-\mathbf{Y}_{s}\|_F^2+\beta\|\mathbf{V}\mathbf{A}_{s}\|_F^2.
    \end{equation}
    ~~~The closed-form solutions of $\mathbf{V}$ can be obtained as:
    \begin{equation}
    \mathbf{V}^*= (\mathbf{C}_{s}\mathbf{C}_{s}^\textrm{T}+\gamma\mathbf{I})^{-1}(\mathbf{C}_{s}\mathbf{Y}_{s}\mathbf{A}_{s}^\textrm{T})(\mathbf{A}_{s}\mathbf{A}_{s}^\textrm{T})^{-1},
    ~~\beta = \alpha\gamma.
    \end{equation}
    ~~~The optimal $\mathbf{D}_{s}$ can be obtained by introducing a variable $\mathbf{R}$:
     \begin{equation}
     \label{Eq.(7)}
    \arg\min_{\mathbf{D}_{s},\mathbf{R}}\|\mathbf{X}_{s}-\mathbf{D}_{s}\mathbf{C}_{s}\|_F^2\\
    ~~\emph{s.t.} ~\mathbf{D}_{s}=\mathbf{R},~ \|r_{i}\|_2^2\leqslant{1}.
    \end{equation}

    And the solution of Eq.~(7) can be obtained by the alternating direction method of multipliers (ADMM) algorithm.

    In each iterative step, $\mathbf{C}_s$ and $\mathbf{V}$ are obtained with closed-form solutions and the optimization of $\mathbf{D}_s$ is obtained with the ADMM algorithm, which converges rapidly. The iterative step stops when the difference between the variations in two adjacent iterations is less than a threshold.

    Once $\mathbf{D}_s^*$ and $\mathbf{V}^*$ are obtained, the compatibility score $s(\mathbf{x},\mathbf{a}_c)$ of a test instance $\mathbf{x}$ over the unseen class $c$ is estimated in the common latent space:
    \begin{equation}
    \label{Eq.(8)}
    s(\mathbf{x},\mathbf{a}_c) = \mathbf{x}^\textrm{T}\mathbf{D}_s^*\mathbf{V}^*\mathbf{a}_{c},
    \end{equation}
    where $\mathbf{a}_{c}$ is the label embedding of the $c$-th unseen class, $(\mathbf{D}_s^{*})^{\mathrm{T}}\mathbf{x}$ is the approximate embedding of visual instance in the latent space, while $\mathbf{V}^*\mathbf{a}_{c}$ is the prototype of unseen class $c$, which is the latent embedding of $\mathbf{a}_{c}$. Thus, ZSL is achieved by resorting to the largest compatibility score with respect to the unseen label embeddings.
    \begin{equation}
    \label{Eq.(8)}
    \emph{c}^*=\max_{c}s(\mathbf{x},\mathbf{a}_c).
    \end{equation}

\subsection{Self-Labeled strategy}

     Like most inductive ZSL approaches, the classification model which is learned only with the labeled seen data will generalize poorly on the unseen data due to that the class distribution of the seen domain is different from that of the unseen domain. To address this domain shift problem, we formulate the prediction step of ZSL in an iterative refining process, in which each iteration alternates between two paradigms, learning-to-predict and predicting-to-learn. With the model learned with the labeled seen data, the labels of the unseen data are previously predicted. This is the first learning-to-predict paradigm. Considering that the instances that have higher compatibility scores are more reliable to be correctly-predicted, it is reasonable to annotate these reliable instances as labeled data for unseen classes. With these feedback reliable instances, the unseen-specific model is retrained for the subsequent prediction step. This is a predicting-to-learn paradigm. Repeat this precess, the domain shift is progressively adapted in a confident way. The remaining problem is how to select reliable instances from unseen data. In this paper, we introduce a simple strategy to select instances from unseen data as labeled data. Specifically, for each unseen class, the test instances can be ranked according to the compatibility scores over their corresponding predicted unseen class. We then set a self-labeled rate $\delta$ to annotate the reliable instances as labeled data. For example, suppose that $n_i$ instances are predicted into the unseen class $y_i$, $[n_i\times{\delta}]$ instances are selected according to their ranking scores to the corresponding class, $[\cdot]$ is the rounding operation. Clearly, the self-labeled strategy is under a transductive setting.


     It should be noted that the self-labeled strategy can be seamlessly integrated into the various existing ZSL approaches. As shown in Fig.~2, the seen data are used for learning a previously classification model for initially predicting the unseen data, and then an iterative strategy is used for refining the learned model. At each iteration, only reliable instances from the unseen data are selected for refining the classification model. As more instances are selected, a powerful specific model is learned for unseen classes.

\subsection{Transductive Self-training Dictionary (TSTD) model}

     By integrating the self-labeled strategy into the previously proposed JEDM, we obtain the final Transductive Self-Training Dictionary (TSTD) model. For the first learning-to-predict paradigm, the class labels of unseen data are previously predicted with the proposed JEDM. And then the classification model is retrained by the unseen data themselves. In each predicting-to-learn paradigm, two baselines are introduced to ensure that the refined model is more suitable for unseen classes. The first one is that the current learned dictionary model $\mathbf{D}_t$ is close to the previously optimal one $\mathbf{D}^*$. Since the previously learned model is used to align different spaces, the currently learned model should refine the previous one by a fine step rather than adjusting with a large range. The other one is that the learned model ensures that the latent embeddings of the self-labeled instances are close to their predicted label prototypes in the latent space. Thus the objective function is defined as follows:

\begin{figure}
    \centering
     \includegraphics[height=6cm,width=8cm]{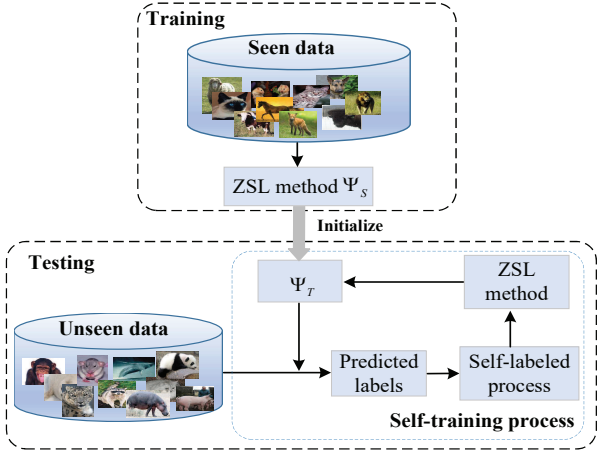}
     \caption{The workflow of our proposed self-training strategy. In the training stage, a ZSL model is trained for $\Psi_{S}$, which is used to initialize the unseen model $\Psi_{T}$. In the testing stage, the labels of the unseen data are predicted by the unseen model $\Psi_{T}$. Then the instances whose labels are reliablely predicted are selected as self-labeled data for refining the unseen model $\Psi_{T}$ with a ZSL method. The self-labeled process stops until all unseen data are selected.}
     \label{Fig.2}
 \end{figure}

    \begin{equation}
    \begin{aligned}
    \{\mathbf{D}_{t}^*,\mathbf{C}^*\}=\arg\min_{\footnotesize{\mathbf{D}_{t},\mathbf{C}}}\|\mathbf{X}-\mathbf{D}_t\mathbf{C}\|_F^2\\
    +\lambda\|\mathbf{V}^*\mathbf{A}-\mathbf{C}\|_F^2
    +\mu\|\mathbf{D}_{t}-\mathbf{D}^*\|_F^2,
    \end{aligned}
    \end{equation}
    where $\mathbf{X}$ is the collected set which contains the selected self-labeled instances, $\mathbf{V}^{*}$ is the previous learned compatibility matrix shared both the seen domain and unseen domain. $\mathbf{D}_t$ is the currently learned dictionary matrix for unseen classes, $\mathbf{C}$ is the latent embeddings of the self-labeled instances and $\mathbf{A}$ is the predicted label embedding matrix that self-labeled instances correspond to. Since each unseen class is associated with a label semantic vector, $\mathbf{A}$ is easily inferred by the predicted class labels. $\lambda$ and $\mu$ are trade-off parameters. In our model, the latent embeddings of the input unseen data are enforced to be close to their corresponding predicted classes' label latent embedding in the common latent space, i.e., $\|\mathbf{V}^*\mathbf{A}-\mathbf{C}\|_F^2$.

    In the following, we design an alternating optimization method to solve Eq.~(10).
     When $\mathbf{D}_{t}$ is fixed, the optimization problem becomes:
    \begin{equation}
    \mathbf{C}^* = \arg\min_{\footnotesize{\mathbf{C}}}\|\mathbf{X}-\mathbf{D}_{t}\mathbf{C}\|_F^2+\lambda\|\mathbf{V}^*\mathbf{A}-\mathbf{C}\|_F^2,
    \end{equation}
    which leads to a closed-form solution:
    \begin{equation}
    \mathbf{C}^* =(\mathbf{D}_{t}^\textrm{T}\mathbf{D}_{t}+\lambda\mathbf{I})^{-1}(\mathbf{D}_{t}^\textrm{T}\mathbf{X}+\lambda\mathbf{V}^*\mathbf{A}).
    \end{equation}

    With the fixed $\mathbf{C}$, the optimal $\mathbf{D}_t^*$ can be easily solved by:
    \begin{equation}
    \mathbf{D}_{t}^*=\arg\min_{\footnotesize{\mathbf{D}_{t}}}\|\mathbf{X}-\mathbf{D}_{t}\mathbf{C}\|_F^2+\mu\|\mathbf{D}_{t}-\mathbf{D}^*\|_F^2.
    \end{equation}

    This is a standard least squares problem, and we have the optimization solution:
    \begin{equation}
    \mathbf{D}_{t}^* =(\mathbf{X}\mathbf{C}^\textrm{T}+\mu\mathbf{D}^*)(\mathbf{C}\mathbf{C}^\textrm{T}+\mu\mathbf{I})^{-1}.
    \end{equation}

    With the currently learned dictionary matrix $\mathbf{D}_t^{*}$, the unseen data are revisited with Eq.~(9). With the latest predicted results, we enlarge the self-labeled rate $\delta$ to incorporate more reliable instances for training. Repeat this refining process until all the unseen data are selected. Specifically, the values of self-labeled rate $\delta$ are successively selected from $\{0.4,0.6,0.8,1\}$ in our experiments. The TSTD process is summarized in Algorithm 1.

 \subsection{Further analysis}

    With the learned dictionary matrix $\mathbf{D}_s^*$ and the compatibility matrix $\mathbf{V}^*$ from the seen data, the unseen instances and the label embeddings of unseen classes can be embedded into a latent space together. We visualize them with t-SNE approach, as illustrated in Fig. 3. We can observe that the projections of most visual instances from the same class are distributed around the corresponding class prototypes in the latent space. It is easy to conclude that the instances that are close to the corresponding class semantic prototypes tend to be classified correctly. In contrast, the instances that are farther away from the corresponding class prototypes tend to be classified into the wrong classes. Thus it is natural to annotate the instances that are close to the corresponding class prototypes as labeled data, which eliminates the issue that no training samples are available for unseen classes. Fixing the prototypes of unseen classes, the embeddings of unseen instances are gradually adjusted by retraining the embedding function with the reliable instances, and thus the domain shift issue in ZSL alleviates. The mechanism of the proposed transductive strategy is borrowing the knowledge from the seen classes to teach unseen data, and then learning a specific model with the unseen data by themselves in a word.



    \begin{figure}
    \centering
     \includegraphics[height=7cm,width=9cm]{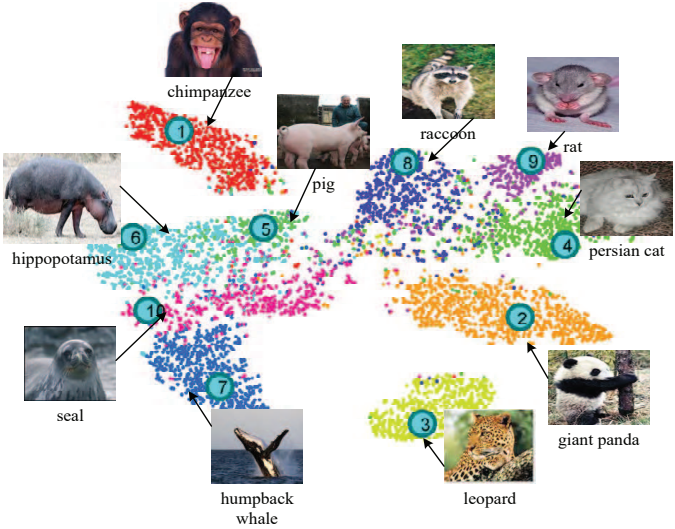}
     \caption{t-SNE visualization of AwA unseen data and the corresponding class attributes embedded in the latent space, where the blue circles denote the embedding prototypes of AwA test classes.}
     \label{Fig.3}
    \end{figure}


    \begin{algorithm}[h]
    \caption{The process of TSTD}
    \label{Algorithm 1}
    \KwIn{\\
    ~~1: The seen domain:\\
     ~~~~~Instance matrix $\mathbf{X}_{s}\in\mathbb{R}^{{p}\times{m}}$,\\
     ~~~~~Ground truth label matrix $\mathbf{Y}_{s}\in{\{-1,1\}}^{{m}\times{M}}$,\\
     ~~~~~Seen label embedding matrix $\mathbf{A}_{s}\in{\mathbb{R}^{{q}\times{M}}}$,\\
     ~~~~~Hyper-parameters $\alpha$,~$\beta$.\\
    ~~2: The unseen domain:\\
     ~~~~~Instance matrix $\mathbf{X}_{t}\in\mathbb{R}^{{p}\times{n}}$,\\
     ~~~~~Unseen label embedding matrix $\mathbf{A}_{t}\in{\mathbb{R}^{{q}\times{N}}}$,\\
     ~~~~~Hyper-parameters $\lambda$,~$\mu$,\\
     ~~~~~Self-labeled rate $\delta\in\{0.4,0.6,0.8,1\}.$
   }
    \KwOut{The predicted class labels of Unseen data.}
    \textbf{Training}:\\
    ~~3:~\textbf{repeat}\\
    ~~4: ~~Update $\mathbf{C}_s$ according to Eq.~(4);\\
    ~~5: ~~Update $\mathbf{V}$ according to Eq.~(6);\\
    ~~6: ~~Update $\mathbf{D}_s$ according to Eq.~(7);\\
    ~~7:~\textbf{until} There is no change to $\mathbf{D}_s$, $\mathbf{C}_s$ and $\mathbf{V}$.\\
    ~~8: \textbf{return} $\mathbf{D}_s^*$ and $\mathbf{V}^*$.\\
    ~~9:~~Fix the compatibility matrix $\mathbf{V}^*$ and initialize the\\
    ~~~~~~unseen dictionary model $\mathbf{D}_{t}^*$ with $\mathbf{D}_{s}^*$ and $\delta$;\\
    ~~10:~\textbf{repeat}\\
    ~~11:~~Predict the unseen data with Eq.~(9);\\
    ~~12:~~\textbf{for} $i=1:N$ \textbf{do}\\
    ~~12:~~~~Rank the instances that are predicted to the\\
    ~~~~~~~~~~unseen class $i$ based on the compatibility scores;\\
    ~~13:~~~~Select the previous $n_i\times\delta$ reliable instances from\\
    ~~~~~~~~~~class $i$ to incorporate the self-labeled set $\mathbf{X}$;\\
    ~~14:~~\textbf{end}\\
    ~~15:~~~~Refer to the label embedding matrix $\mathbf{A}$ according\\
    ~~~~~~~~~~to the predicted labels of the selected $\mathbf{X}$;\\
    ~~16:~~~~Update $\mathbf{C}$ according to Eq.~(12);\\
    ~~17:~~~~Update $\mathbf{D}_t^{*}$ according to Eq.~(14);\\
    ~~18:~~~~Enlarge the self-labeled rate $\delta$.\\
    ~~19:~\textbf{until} All the unseen instances are selected.
    \end{algorithm}


\subsection{Complexity Analysis}
    In this section, we analyze the computational complexity of TSTD and the convergence of the proposed JEDM separately.
\par\setlength\parindent{1em}
    \textbf{Computational Complexity.} In the training phase of JEDM, $\mathbf{D}_{s}$, $\mathbf{C}_{s}$ and $\mathbf{V}$ are updated alternatively. In each iteration, the time complexities of updating $\mathbf{C}_{s}$ and $\mathbf{V}$ in Eq.~(4) and Eq.~(6) are $O(mpd + d^{3} + d^{2}m)$ and $O(dmMq + q^{3} + q^{2}dm + d^{3} + d^{2}qM)$, respectively. As for the optimization of updating $\mathbf{D}_{s}$, the time cost is about $O(K(pmd + d^{3} + d^{2}p + p^{2}d))$, where $K$ is the iteration number in ADMM algorithm. We have experimentally found that the ADMM algorithm converges with less than 20 iterations. In the domain adaptation phase of TSTD, $\mathbf{C}$ and $\mathbf{D}_{t}$ are also updated alternatively. In each iteration, the time complexities of updating $\mathbf{C}$ and $\mathbf{D}_{t}$ are $O(npd\delta + d^{3} + d^{2}n\delta)$ and $O(dpn\delta + d^{3} + d^{2}p)$, respectively. Given that $M\ll{d}$, $q\ll{d}$, and $m$, $p$, $n$, $d$ are in the same order of magnitude and our algorithm converges with a few iterations, the over time cost of our algorithm is $O(d^{3})$. It is worth noting that the dominant operation of our algorithm is matrix multiplication, which can greatly accelerate the training process.


    \textbf{Convergence.} We conduct empirical study on the convergence property using Animal with Attribute (AwA) with attributes as label semantic vectors.
    We set the hype-parameters $\alpha$ and $\beta$ both as 0.1.
    The train/test split provided by the dataset is used accordingly.
    As Fig. 4 shows, the cost function of JEDM descends dramatically and converges with only 10 iterations, which clearly indicates the efficiency of the proposed JEDM.\\
\begin{figure}
    \centering
     \includegraphics[height=6cm,width=7cm]{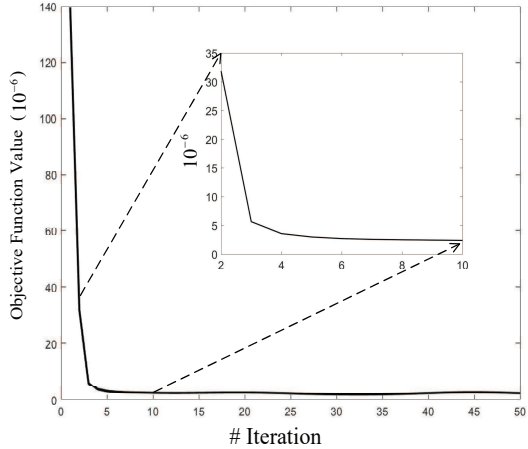}
     \caption{The convergence curve of JEDM on the AwA dataset with attributes as label embeddings.}
     \label{Fig.3}
 \end{figure}


\section{Experiments}
In this section, we do a set of experiments to demonstrate the superiority of the proposed approaches. Firstly, we detail the datasets and settings for the experiments, and then compare the proposed JEDM with the state-of-the-art inductive ZSL approaches. Then, the effectiveness of the proposed self-training strategy is evaluated, followed by the comparison results about TSTD and the state-of-the-art transductive ZSL approaches.
\subsection{Datasets and Settings}

    \textbf{Datasets.} To evaluate the effectiveness of the proposed approaches, we conduct extensive experiments on three benchmark datasets. (a). \textbf{Animal with Attribute (AwA)} \cite{IEEEcvpr09:Lampert} consists of 30,475 animal images from 50 different classes, and each class is associated with a 85-dimensional attribute vector. (b). \textbf{Caltech-UCSD Bird2011 (CUB)} \cite{Technical11:Wah} is a fine-grained dataset which contains~11,788 images from 200 bird subspecies, and a 312-dimensional attribute vector is provided for each class. (c). \textbf{SUN Attribute} \cite{IJCV14:Patterson} contains 717 scene categories annotated by 102 attributes, and each class has 20 images. For the seen/unseen class split, we use the standard 40/10 split setting for AwA dataset \cite{IEEEcvpr09:Lampert}. For CUB dataset, we follow the same 150/50 split in \cite{IEEEcvpr15:Akata}. And for SUN dataset, we use 707 classes as the seen domain and 10 classes as the unseen domain, the same as that in \cite{IEEEcvpr16:Zhang}. The statistics for the three datasets are shown in TABLE \uppercase\expandafter{\romannumeral2}.

    \begin{table}
    \label{Table.1}
    \caption{\upshape The statistics of three benchmark datasets}
    \centering
    \begin{tabular}{c|c|c|c}
    \hline
    Dataset & instances & attribute & seen/unseen classes\\
    \hline
    \hline
    AwA & 30,475 & 85 & 40/10\\
    \hline
    CUB & 11,788 & 312 & 150/50\\
    \hline
    SUN & 14,340 & 102 & 707/10\\
    \hline
    \end{tabular}
    \end{table}

    \textbf{Visual representation.} In our experiments, we use the vgg-verydeep-19 (denoted as VGG for short) features provided by those datasets for representing the visual instances.

    \textbf{Label semantic embedding.} In this paper, we explore the visual attributes and  word vectors as label embedding space for AwA and CUB datasets. Meanwhile, only visual attributes are used for SUN dataset to be comparable with the existing practices in the literature.

    Besides, there are four hyper-parameters $\alpha$, $\beta$, $\lambda$ and $\mu$ in our proposed TSTD, $\alpha$ and $\beta$ are two parameters in the JEDM and $\lambda$ and $\mu$ are in the refining model. We select their best values with a 5-fold cross-validation (CV) strategy, where 20\% of the seen classes are held out for validation and the remaining for training. Once the parameters are fixed, all seen classes are then trained together for the final model. All the parameters are selected from $\{0.01,0.1,1,10,100\}$. In all the experiments, the classification performances are evaluated with the average per-class top-1 accuracy. The average running time of our Matlab implementation is about 0.01ms per image on a desktop with an Intel Core i7-4790K processor and 32G RAM.

\subsection{Comparative results of JEDM}

    In order to evaluate the effectiveness of the proposed JEDM, we conduct two experiments according to the types of label embedding space.

    We first take attributes as semantic vectors for classes. In this experiment, six state-of-the-art approaches are selected for comparison. For descriptive convenience, they are respectively referred to as DAP (Direct Attribute Prediction \cite{IEEEpami:Lampert}), SJE (Structal Joint Embedding \cite{IEEEcvpr15:Akata}), LatEm (Latent Embeddings \cite{IEEEcvpr16:Xian}), ESZSL (Embarrassing Simple Zero-Shot Leaning \cite{IEEEnips15:Romera-Paredes}), SC (Synthesized Classifiers \cite{IEEEcvpr16:Changpinyo}) and JLSE (Joint Latent Similarity Embedding \cite{IEEEcvpr16:Zhang}). These selected competing methods are all inductive approaches.

     The results of the comparative methods are all from the original papers except \cite{IEEEnips15:Romera-Paredes}, which is obtained with the published codes under the same setting as ours. The results are summarized in TABLE \uppercase\expandafter{\romannumeral3}, where `-' indicates that these methods were not tested on the datasets in their original work.

    From TABLE \uppercase\expandafter{\romannumeral3}, we can observe that JEDM is comparable with the state-of-the-art approaches.
    More specifically, in the AwA dataset, JEDM achieves an improvement of 19.0\% against the baseline method DAP \cite{IEEEpami:Lampert} and beats the other competitors expect for JLSE \cite{IEEEcvpr16:Zhang}, which projects both modalities into different latent spaces. It is a more complicated model. For CUB dataset, our approach works better than others except for \cite{IEEEcvpr16:Changpinyo} and \cite{IEEEcvpr15:Akata}. \cite{IEEEcvpr16:Changpinyo} tackles ZSL with exploiting the manifold structure to align the semantic space, which behaves robust for the fine-grained dataset. While \cite{IEEEcvpr15:Akata} takes a more powerful visual feature as the input, which attributes to the fact that \cite{IEEEcvpr15:Akata} works better than JEDM. Since the SUN dataset is less popular than the above two, only three recent approaches are selected for comparison. From the results, we can find that the proposed JEDM outperforms the previously published approaches by a large margin. Specifically, it outperforms DAP \cite{IEEEcvpr09:Lampert}, ESZSL \cite{IEEEnips15:Romera-Paredes} and \cite{IEEEcvpr16:Zhang} in 14\%, 4\% and 3.2\% gains, respectively.
      Besides, it is found that classification performances of JEDM outperform that of ESZSL, which is similar to the proposed JEDM. The most difference between our method and ESZSL is that JEDM projects the visual features into a more discriminative latent space with a dictionary framework, while ESZSL uses the visual feature as input directly and designs an elaborated regularizer. The comparative results demonstrate the effectiveness of the dictionary representation.
\begin{table}
  \label{Table.1}
  \caption{\upshape Comparison results of different approaches on different datasets with attributes (in \%). Notations: `F': visual features; `V': VGG feature; `G': GoogleNet feature, $\S$ indicates the method of which the classification performances are obtained by ourselves. We report the best performance after tuning the parameters in their models.}
  \centering
  \begin{tabular}{c|c|c|c|c}
  \hline
  Method & F & AwA & CUB & SUN\\
  \hline
  \hline
  DAP \cite{IEEEpami:Lampert} & V & 57.5 & - & 72.0\\
  SJE \cite{IEEEcvpr15:Akata} & G  & 66.7 & 50.1 & -\\
  LatEm \cite{IEEEcvpr16:Xian} & G & 71.9 & 45.5 & -\\
  ESZSL \cite{IEEEnips15:Romera-Paredes}$\S$ & V & 75.3 & 46.8 & 82.0\\
  Changpinyo \emph{et al.}\cite{IEEEcvpr16:Changpinyo} & G & 72.9 & \textbf{54.7} & -\\
  JLSE \cite{IEEEcvpr16:Zhang} & V & \textbf{80.5} & 42.1 & 82.8\\
  \hline
  JEMD &V & 76.5 & 47.6 & \textbf{86.0}\\
  \hline
  \end{tabular}
\end{table}

    In the second experiment, the word vector space is taken as the label embedding space. Thanks to the recent advances in unsupervised neural language modeling \cite{IEEEnips13:Mikolov} \cite{IEEEemnl14:Pennington}, each word in a text corpus can be effectively embedded in a textual semantic space, where each word is represented as a semantic multi-dimensional vector. Specifically, we use word2vector model \cite{IEEEnips13:Mikolov} to train a skip-gram language model on the latest Wikipedia corpus to extract 1000-dimensional word vector for each class from AwA and CUB datasets.
    Five wordvector-based approaches are selected for comparison, as illustrated in TABLE \uppercase\expandafter{\romannumeral4}. From the results, we can find that JEDM has an impressive improvement in AwA dataset. Specially, JEDM outperforms CCA \cite{ACL14:Lazaridou}, SJE \cite{IEEEcvpr15:Akata}, LatEm \cite{IEEEcvpr16:Xian} and ESZSL \cite{IEEEnips15:Romera-Paredes} in 5.9\%, 20.3\%, 10.4\% and 4.1\% gains, respectively. Meanwhile, it also achieves a competitive result in CUB dataset, which is only 0.9\% lower than that of the previous best reported LatEm \cite{IEEEcvpr16:Xian}.

\begin{table}
  \label{Table.2}
   \caption{\upshape Comparison results on AwA and CUB datasets with word vectors (in \%). Notations: `V': VGG feature;
  `G': GoogleNet feature. $\S$ indicates the methods that the classification performances are obtained by ourselves.}
  \centering
  \begin{tabular}{c|c|c|c}
  \hline
  Method & F & AwA & CUB\\
  \hline
  \hline
  CCA \cite{ACL14:Lazaridou}$\S$ & V & 65.6 & 30.4\\
  SJE \cite{IEEEcvpr15:Akata} & G  & 51.2 & 28.4\\
  LatEm \cite{IEEEcvpr16:Xian} & G & 61.1 & \textbf{31.8}\\
  ESZSL \cite{IEEEnips15:Romera-Paredes}$\S$& V & 67.4 & 30.4\\
  \hline
  JEMD &V & \textbf{71.5} & 30.9\\
  \hline
  \end{tabular}
\end{table}

\subsection{Evaluation of self-training strategy}
    In this section, we conduct a set of experiments on AwA and CUB datasets to demonstrate the generality and the effectiveness of the proposed self-training strategy. In specific, two typical ZSL approaches are selected for being integrated with the self-training strategy. These approaches are CCA and ESZSL, both of which have a closed-form solution. For descriptive convenience, we add a postfix -ST to the name of the approaches for representing the corresponding approaches with the self-training strategy. Specifically, the approach that JEDM integrates self-training strategy is called TSTD in this paper. In implementation, the baselines introduced in TSTD are also suitable for CCA-ST and ESZSL-ST. The comparative results are provided in TABLE~\uppercase\expandafter{\romannumeral5}.

    From the results, we can observe that the proposed transductive self-training strategy can not only improve the performance of the proposed JEDM with a large margin, but also boost other approaches substantially on different datasets with different semantic vectors. Specifically, on AwA dataset, the transductive self-training strategy helps JEDM improve 13.8\% and 19.7\% in gains with attribute and word vector as label embedding space, respectively. It should be noted that TSTD achieves 91.2\% classification accuracy on AwA dataset with word vector as semantic space, which is even better than those attribute-based approaches. In contrast to AwA dataset, the improvement range of the transductive self-training strategy is smaller on CUB dataset.
    The reason is that the CUB dataset is a fine-grained dataset and its classification performance of JEDM is much lower than that of AwA dataset, such that the self-labeled set contains many fake instances that may spoil the classification model.
    Even so, the proposed transductive self-training strategy helps JEDM improve 10.6\% and 3.0\% absolute percentage points with visual attribute and word vector as semantic space, respectively.
 \begin{table}
    \label{Table.5}
    \caption{\upshape Evaluations of self-training strategy on AwA and CUB datasets, `M-ST' denotes the method `M' with self-training strategy, A and W are short for visual attribute and word vector, respectively}
    \centering
    \begin{tabular}{c|c|c|c|c}
    \hline
    \multirow{2}{*}{Method} &\multicolumn{2}{c|}{AwA} &\multicolumn{2}{c}{CUB}\\
    \cline{2-5}
     &A &W &A &W\\
    \hline
    CCA \cite{ACL14:Lazaridou} &72.5 &65.6 &46.2 &30.4\\
    ESZSL \cite{IEEEnips15:Romera-Paredes} &75.3 &67.4 &46.8 &30.4\\
    JEDM&76.5 &71.5 &47.6 &30.9\\
    \hline
    CCA-ST &85.6 &79.9 &53.8 &32.6\\
    ESZSL-ST &87.7 &83.8 &56.4 &32.4\\
    TSTD &\textbf{90.3} &\textbf{91.2}& \textbf{58.2} &\textbf{33.9}\\
    \hline
    \end{tabular}
\end{table}

\subsection{Comparison results of TSTD}
   We also compare our TSTD approach with the state-of-the-art transductive ZSL approaches. TABLE \uppercase\expandafter{\romannumeral6} shows the comparison results. We can observe that the proposed TSTD has an overwhelming superiority to the competitors. Specifically, the proposed domain adaptation strategy on JEMD model achieves 90.3\% classification accuracy on AwA dataset with visual attribute, which outperforms \cite{IEEEpami15:Fu}, \cite{IEEEaaai16:Guo}, \cite{IEEEiccv15:Kodirov} and \cite{arXiv16:Wang} in 9.8\%, 11.8\%, 14.7\% and 2.4\% gains, respectively. On CUB dataset, it achieves 58.2\% classification accuracy, which improves 10.3\%, 17.6\% and 4.7\% over \cite{IEEEpami15:Fu}, \cite{IEEEiccv15:Kodirov} and \cite{arXiv16:Wang}, respectively. Specifically, TMV-HLP and SMS are two transductive methods that integrate the seen data and unseen data together for training a general model for all classes. And \cite{arXiv16:Wang} explores the label information of unseen data with an unsupervised cluster-based approach. However, \cite{IEEEiccv15:Kodirov} and our self-training strategy focus on re-training a suitable model for unseen classes. The main difference between these two strategies is that \cite{IEEEiccv15:Kodirov} uses an unsupervised model to exploit the structure information of the unseen domain while ours relies on a bootstrapping-based model updating over highly reliable instances to progressively reinforce the classification capacity.
\begin{table}
  \label{Table.2}
   \caption{\upshape Comparison results on different transductive approaches (in \%), where visual attribute is adopted as the semantic space.}
  \centering
  \begin{tabular}{c|c|c}
  \hline
  Method  & AwA & CUB\\
  \hline
  \hline
  TMV-HLP \cite{IEEEpami15:Fu}  & 80.5 & 47.9\\
  SMS \cite{IEEEaaai16:Guo} & 78.5 & -\\
  Kodirov \emph{et al.} \cite{IEEEiccv15:Kodirov} & 75.6 & 40.6\\
  Wang \emph{et al.} \cite{arXiv16:Wang} &87.9 & 53.5 \\
  \hline
  TSTD &\textbf{90.3}  & \textbf{58.2}  \\
  \hline
  \end{tabular}

\end{table}

\subsection{Evaluation of the self-labeled rate}
 We next conduct a set of experiments to evaluate the influences of self-labeled rate $\delta$ to the maturity of the learned model. As illustrated in Fig.~5, we can observe that the performances increase steadily with the increase of $\delta$ and achieve their peaks when $\delta=0.8$ on AwA dataset with different types of label semantic embeddings. This indicates that with the increase of $\delta$, more correct self-labeled instances are selected for refining the classification model, thus the classification capacity is progressively reinforced. In contrary, on CUB dataset, the performances achieve their peaks when $\delta=0.6$  and $\delta=0.4$ with attributes and word vector, respectively. And the performances decrease with increase of $\delta$. This is due to the classification performances on CUB unseen data with the learned model are poor (47.6\% and 30.9\% with attribute and word vector respectively), and thus with increase of $\delta$, more false instances are selected as self-labeled data, which may spoil the learned model. The curves of AwA dataset in Fig.~5 (a) also verify this explanation.
 \begin{figure}[h]
    \subfigure[AwA]{
        \label{fig5_a}
\includegraphics[height=2.8cm,width=1.6in]{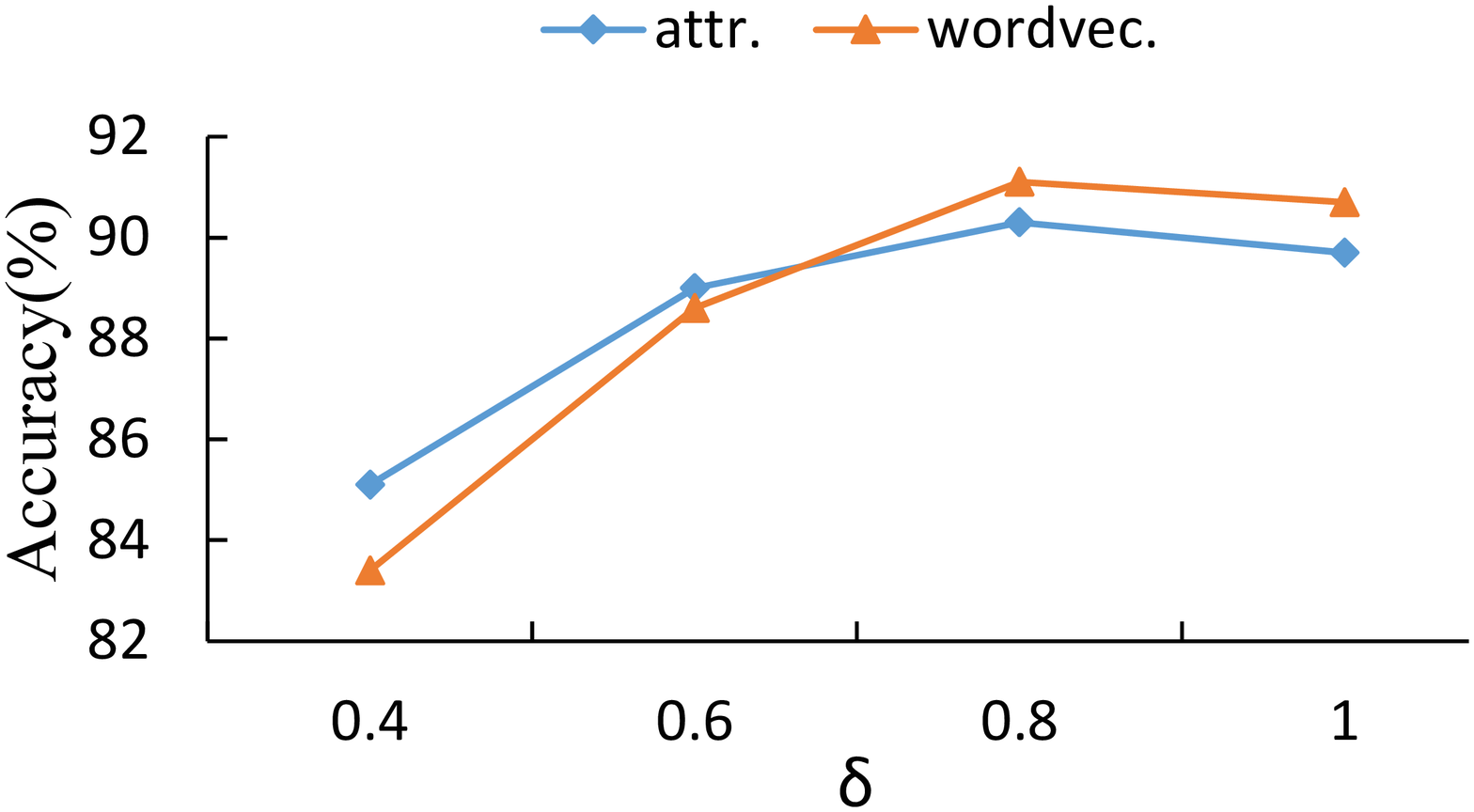}}
    \hspace{0.01in}
    \subfigure[CUB]{
        \label{fig5_b}
\includegraphics[height=2.8cm,width=1.6in]{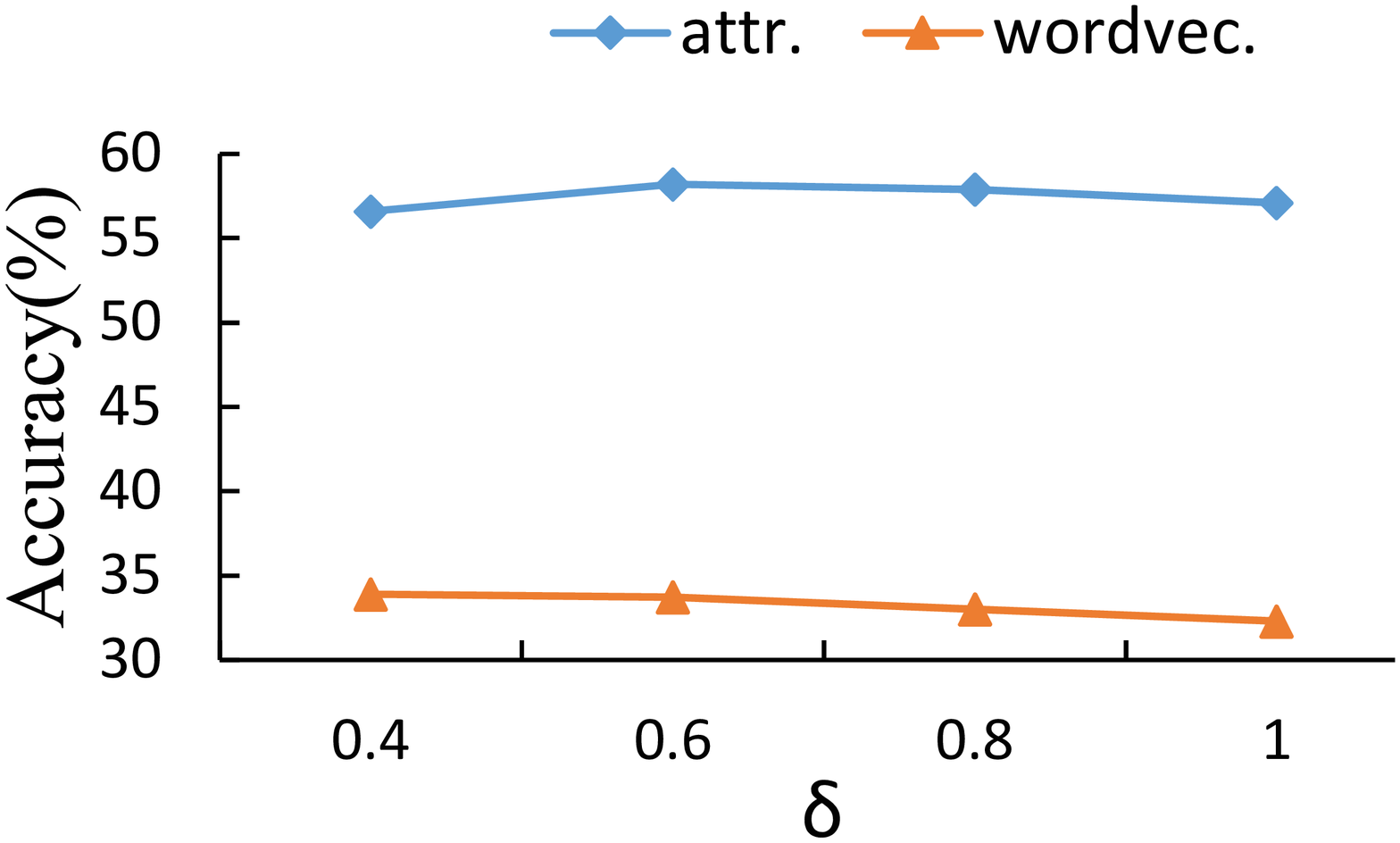}}
    \caption{The classification performances with different rates $\delta$ for TSTD on AwA and CUB datasets, respectively.}
    \label{fig4}
\end{figure}

\section{Conclusions}

    In this paper, we proposed a bidirectional mapping based scheme to address ZSL. It formulates the semantic interactions between image feature space and label embedding space in a general dictionary model by simultaneously projecting the image features and label embeddings into a common latent space. The experimental results demonstrated that the proposed approach achieves the state-of-the-art performance on three benchmark datasets. To alleviate the domain shift problem in ZSL, we further proposed a transductive learning framework that formulates ZSL in two paradigms, where the labeled seen data are used to transfer the knowledge to unseen data, and the unlabel unseen data are used to gradually learn a more powerful model by themselves. In this way, the classification capacity is progressively reinforced through bootstrapping-based model updating over highly reliable unseen instances. The experimental results demonstrated that the proposed transductive strategy improves the classification performance of the existing inductive methods with a large margin. Compared with the state-of-the-art methods, our transductive approach outperforms the runner-up method on AwA and CUB datasets with 2.4\% and 4.7\% improvements, respectively.

\ifCLASSOPTIONcaptionsoff
  \newpage
\fi

\end{document}